\pdfoutput=1

\documentclass[11pt]{article}

\usepackage[final]{acl}
\usepackage{times}
\usepackage{latexsym}

\usepackage[T1]{fontenc}

\usepackage[utf8]{inputenc}

\usepackage{microtype}

\usepackage{inconsolata}

\usepackage[utf8]{inputenc} 
\usepackage[T1]{fontenc}    
\usepackage{hyperref}       
\usepackage{url}            
\usepackage{booktabs}       
\usepackage{amsfonts}       
\usepackage{nicefrac}       
\usepackage{microtype}      

\usepackage{graphicx}
\usepackage{algorithm}
\usepackage{algorithmic}
\usepackage{arydshln}
\usepackage{comment}
\usepackage{multirow} 
\usepackage{booktabs}
\usepackage{wrapfig}
\usepackage{array}
\usepackage{adjustbox}
\usepackage{amsmath}
\usepackage{pifont}  

\usepackage{graphicx}
\usepackage{caption}
\usepackage{subcaption}

\makeatletter
\def\@fnsymbol#1{%
  \ifcase#1\or
    \ding{168}
  \or
    \ding{169}
  \else
    \@ctrerr
  \fi
}
\makeatother

\definecolor{ultramarine}{RGB}{0,32,96}
\definecolor{wrongultramarine}{RGB}{0.07, 0.04, 0.56}
\definecolor{danshuhong}{RGB}{234, 107, 102}
\definecolor{purpleforus}{RGB}{181, 115, 157}
\definecolor{yalv}{RGB}{150, 194, 78}
\definecolor{molv}{RGB}{103, 171, 159}
\definecolor{orangelight}{RGB}{255, 179, 102}
\definecolor{yellowdark}{RGB}{239,236,0}
\definecolor{bluedark}{RGB}{126,166,224}
\definecolor{newpurple}{RGB}{181,115,157}
\definecolor{lightpink}{RGB}{255,208,215}
\definecolor{lightblue}{RGB}{164, 221, 237}

\hypersetup{
    colorlinks=true,
    linkcolor=red,  
    urlcolor=magenta,
    filecolor=magenta,
    citecolor=blue,  
    pdftitle={Overleaf Example},
    pdfpagemode=FullScreen,
}

\title{Arg-LLaDA: Argument Summarization via Large Language Diffusion Models and Sufficiency-Aware Refinement}



\author{%
  Hao Li$^{1}$ \quad  
  {\bf Yizheng Sun$^1$} \quad 
  Viktor Schlegel$^{1, 2}$ \\
  {\bf Kailai Yang$^1$} \quad 
  {\bf Riza Batista-Navarro$^1$} \quad  
  {\bf Goran Nenadic$^1$}  \\
  $^1$ The University of Manchester, UK \quad \\
  $^2$ Imperial Global Singapore, Imperial College London, Singapore \\
  \small{
   \textbf{Correspondence:} kailai.yang@manchester.ac.uk
 }
}

\begin{document}
\maketitle
\begin{abstract} 
Argument summarization aims to generate concise, structured representations of complex, multi-perspective debates. While recent work has advanced the identification and clustering of argumentative components, the generation stage remains underexplored. Existing approaches typically rely on single-pass generation, offering limited support for factual correction or structural refinement. To address this gap, we introduce Arg‑LLaDA, a novel large language diffusion framework that iteratively improves summaries via sufficiency-guided remasking and regeneration. Our method combines a flexible masking controller with a sufficiency-checking module to identify and revise unsupported, redundant, or incomplete spans—yielding more faithful, concise, and coherent outputs. Empirical results on two benchmark datasets demonstrate that Arg‑LLaDA surpasses state-of-the-art baselines in 7 out of 10 automatic evaluation metrics. In addition, human evaluations reveal substantial improvements across core dimensions, coverage, faithfulness, and conciseness,validating the effectiveness of our iterative, sufficiency-aware generation strategy.
\end{abstract}

\section{Introduction}

Argument summarization aiming to distill salient points from multi-perspective debates (known as \textit{Key Point
Analysis (KPA)} \citep{DBLP:conf/acl/Bar-HaimEFKLS20,DBLP:conf/emnlp/Bar-HaimKEFLS20}) or argumentative texts into concise, faithful, and well-structured summaries \citep{DBLP:conf/acl/0074WSBMZZWHLN24} (Shown in Table \ref{tab:arg_example}). This task is particularly important for applications in civic discourse analysis \citep{DBLP:conf/acl/VecchiFJL20}, legal decision support \citep{DBLP:journals/ail/HabernalFRBGDB24}, and educational technology \citep{DBLP:journals/eis/GrljevicBK22}, where it is essential to capture the reasoning structures and factual bases behind opposing claims. Current pipeline can be decomposed into two critical stages: \emph{(1)} identifying and collecting the core argumentative components—namely, claims and their supporting or opposing evidence and \emph{(2)} generating high-quality summaries based on this structured input \cite{DBLP:conf/eacl/MeerVJM24}. 

\begin{table*}[t]
\centering
\renewcommand{\arraystretch}{1.3}
\begin{tabular}{|p{15.5cm}|}
\toprule
\textbf{Input} \\
\midrule
\textbf{(Topic)} Routine child vaccinations should be mandatory \\

\textbf{(Stance)} Oppose \\

\textbf{(Claim 1)} Vaccines or their side effects may be dangerous \\

\textbf{(Evidence 1)} Rotashield was withdrawn after being linked to bowel obstruction \\

\textbf{(Evidence 2)} CDC reports rare risks like pneumonia (chickenpox vaccine) and Guillain-Barré Syndrome (flu vaccine) \\

\textbf{(Claim 2)} Mandatory vaccination violates basic rights \\

\textbf{(Evidence 1)} The First Amendment protects religious freedom \\

\textbf{(Evidence 2)} Compulsory vaccination interferes with bodily integrity, violating international human rights conventions \\
\midrule
\textbf{Summary} \\
\midrule
Mandatory child vaccinations raise health and ethical concerns. Historical cases like Rotashield and CDC findings highlight medical risks. Moreover, opponents argue that such mandates may infringe on personal freedoms and human rights, suggesting vaccination should remain a personal choice guided by awareness rather than coercion. \\
\bottomrule
\end{tabular}
\caption{Example mapping from structured arguments to generated summary. The model effectively integrates two claims and four supporting evidences into a coherent output.}
\label{tab:arg_example}
\end{table*}

Recent advances in argument summarization have concentrated on what we refer to as Stage 1, which either involves: \emph{(i)} constructing argumentative components from a fixed, pre‑collected corpus; or  \emph{(ii)} retrieving such components dynamically from the open Web. Some studies target the former, automatically identifying argumentative elements—such as claims, premises, and stance—via supervised learning methods \citep{DBLP:conf/argmining/KapadnisPPMN21} or prompt‑based large language models (LLMs) \citep{DBLP:conf/eacl/TangZD24,DBLP:conf/acl/Tang0DC24,DBLP:journals/corr/abs-2404-18371}. Other work further groups semantically and functionally similar arguments into representative key points \citep{DBLP:conf/argmining/PhanNND21,DBLP:conf/argmining/ReimerLHA21}, employing techniques such as contrastive learning \citep{DBLP:conf/argmining/AlshomaryGSHSCP21}, graph‑based \citep{DBLP:conf/naacl/LiJHXCH24}, and density‑based clustering \citep{DBLP:conf/acl/LiSBN23,DBLP:conf/naacl/Khosravani0T24}. Meanwhile, another strand of work emphasizes dynamic retrieval of contextual evidence from external sources such as online documents or knowledge bases: models query these sources at inference time to retrieve supporting or elaborating context for argumentative components \citep{DBLP:journals/ijon/ShanL25}. This retrieval‑augmented generation (RAG) paradigm enables LLMs to ground their outputs in external content, enhancing factuality and currency without continually training the model.

However, Stage 2 remains inadequately investigated. Firstly, most existing approaches rely on end-to-end fine-tuned models \citep{eden2023welcome,DBLP:conf/acl/Bar-HaimEKFS20} or prompt-based LLMs that generate summaries \citep{DBLP:conf/acl/Tang0DC24,DBLP:journals/corr/abs-2503-00847} in an auto-regressive manner, providing no inherent mechanism for correction. Once the summary is generated, post-hoc editing becomes cumbersome and unsystematic. This greedy decoding strategy often favors highest-probability tokens at each step, resulting in factual inaccuracies or hallucinations \citep{DBLP:journals/corr/abs-2309-16459}. Secondly, unlike text summarization, argument summarization demands conciseness, covers different aspects or facets without redundancy, with a clear argumentative structure \citep{li2025large}. Existing models rarely guarantee the preservation of argument roles, such as claim-premise relations, or tightly control thematic redundancy \citep{DBLP:conf/acl/0074WSBMZZWHLN24, DBLP:journals/corr/abs-2304-07666, DBLP:conf/acl/ChengBHYZS22}. As a result, these methods often compromise narrative coherence and structural fidelity, rendering them ill-suited for producing argument summaries that meet the required standards.

To address these challenges, we introduce \textbf{Arg‑LLaDA}, a novel large language diffusion framework specifically tailored for argument summarization. Rather than generating summaries in a one-shot, greedy fashion, \textbf{Arg‑LLaDA} employs an iterative remasking and regeneration strategy designed to incrementally refine outputs through multiple editing cycles.  Central to our method is a novel enhanced span-infilling paradigm \citep{DBLP:journals/corr/abs-2502-09992} with a flexible mask-remask controller, which allows for multiple masking strategies—such as sentence-level masking, thereby enabling precise and context-aware rewriting of deficient summary segments while preserving global coherence.  Additionally, we incorporate a sufficiency-checking module, which detects unsupported, redundant, or logically incomplete spans based on the principle of conclusion reconstructability. We validate the effectiveness of our design on the ASE and ArgKP benchmarks. Arg‑LLaDA outperforms state-of-the-art LLM baselines such as Gemini 2.5 and DeepSeek-R1, achieving relative improvements with 7 of 10 metrics best. Human evaluation further validates these gains, with Arg‑LLaDA improving coverage, faithfulness, and conciseness by up to 5.0\% compared to baselines. Ablation studies demonstrate that the iterative refinement and sufficiency diagnosis modules contribute orthogonally to model improvements, showing their independent effectiveness.

In summary, our key contributions are as follows:  \emph{(1)} \textbf{Arg‑LLaDA}, a novel self-refinement model that improves argument summaries via sufficiency-guided remasking and iterative generation. \emph{(2)} We empirically validate our approach on benchmark datasets, demonstrating significant improvements in content fidelity, conciseness, and argumentative structure over prior baselines.

\section{Related work}

\textbf{Argument Summarization.} Traditional approaches often relied on extractive methods, such as clustering-based selection \citep{DBLP:journals/corr/abs-1711-00092, DBLP:conf/acl/ReimersSBDSG19, DBLP:conf/emnlp/AjjourAWS19}, which group similar arguments without producing fluent summaries. Later work reformulated the task as one of claim generation \citep{DBLP:conf/naacl/WangL16} or aspect-controlled generation \citep{DBLP:conf/naacl/SchillerDG21}, enabling abstractive summaries that better capture the underlying stance. A prominent research line is \textit{Key Point Analysis (KPA)} \citep{DBLP:conf/acl/Bar-HaimEFKLS20,DBLP:conf/emnlp/Bar-HaimKEFLS20}, which identifies representative key points across a collection of arguments. Follow-up work enhances this pipeline through unsupervised clustering \citep{DBLP:conf/argmining/AlshomaryGSHSCP21}, generation-based key point synthesis \citep{DBLP:conf/acl/LiSBN23}, and extractive key point selection with structure-aware scoring \citep{DBLP:conf/naacl/Khosravani0T24}. Recent studies explore end-to-end generation models that condition on structured argumentative inputs \citep{DBLP:conf/nips/ChenL0022} or integrate content selection via reinforcement learning \citep{DBLP:conf/acl/RoitFSACDGGHKMG23}. 

\paragraph{Diffusion Language Models.} Recent work has increasingly focused on diffusion-based modeling, encompassing both continuous and discrete approaches \citep{DBLP:journals/corr/abs-2503-02445}. Notably, masked diffusion models, a type of discrete diffusion technique, have emerged as the leading variant in terms of performance. Theoretical studies  have provided rigorous foundations for these models and verified their competitiveness with autoregressive architectures at the GPT-2 scale \citep{DBLP:conf/nips/ShiHWDT24, DBLP:conf/iclr/OuNXZSLL25, DBLP:journals/corr/abs-2402-03300, DBLP:journals/corr/abs-2506-07584}. Building on this progress, LLaDA \citep{DBLP:journals/corr/abs-2502-09992} scaled masked diffusion modeling to 8 billion parameters—becoming the first diffusion-based LLM to rival models like LLaMA 3 \citep{DBLP:journals/corr/abs-2407-21783} in a variety of downstream tasks. Its multimodal variant, LLaDA‑V \cite{you2025llada}, outperforms the LLMs baseline across several multimodal benchmarks—indicating potential inherent advantages of the masked-diffusion framework in broader, multimodal contexts.  

\section{Methodology}

\begin{figure*}[t]
    \centering
    \includegraphics[width=\textwidth]{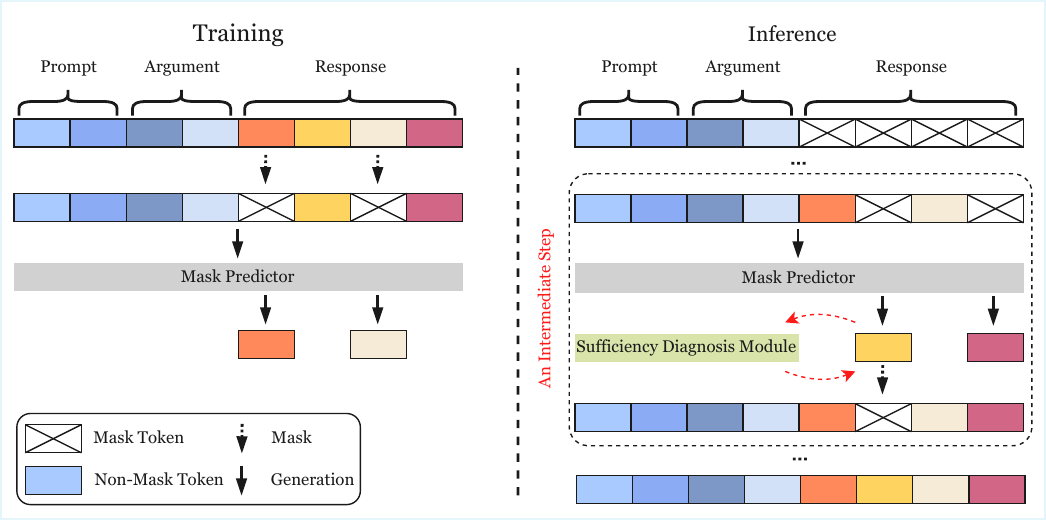}
    \caption{
    \textbf{Training and inference process of Arg-LLaDA.}
    The model iteratively generate the output through a masked denoising diffusion process.
    At each timestep, the sufficiency diagnosis module assigns token-level sufficiency scores based on the claim-evidence context, guiding a selective masking controller to focus regeneration on unsupported or redundant spans.
    This allows Arg-LLaDA to perform semantically grounded and targeted refinement toward factually faithful and concise argument summaries.
    }
    \label{fig:main}
\end{figure*}

In this section, we introduce \textbf{Arg-LLaDA}, a diffusion-based framework for argument summarization. We first formalize the task (Section~\ref{problem}), then describe an iterative masked denoising process with sufficiency-guided masking (Section~\ref{diffusion}). To support targeted refinement, we develop a sufficiency diagnosis module using both prompting and supervised classification (Section~\ref{sufficiency}). Finally, we outline training objectives and implementation details (Section~\ref{training}).

\subsection{Problem Definition}
\label{problem}

We study the task of \emph{argument summarization}, where the input is a collection of $N$ claim--evidence pairs:
\begin{equation}
\{(c_i, E_i)\}_{i=1}^{N}
\end{equation}
with each claim \(c_i\) supported by its corresponding set of evidence \(E_i\). Our goal is to generate a summary \(S\) satisfying the following criteria: \emph{(i)} \textbf{Comprehensiveness}: The summary should include all relevant claims from the input. \emph{(ii)} \textbf{Conciseness}: The output must avoid redundancy and unnecessary verbosity. \emph{(iii)} \textbf{Faithfulness}: All summarized claims and conclusions must be grounded in the provided evidence. We formalize the problem as finding a summary
\begin{equation}
S^* = \arg\max_{S}\; \text{Quality}(S \mid \{(c_i, E_i)\})
\end{equation}
under the above desiderata. 

\subsection{Sufficiency Diagnosis Module}
\label{sufficiency}

To implement the global component of inference-time masking, we introduce a sufficiency diagnosis module that evaluates whether generated spans are adequately grounded in the input arguments. 
Here, a summary span is defined at the sentence level (or clause-level segmentation for long sentences) to ensure coherent semantic units. 
A span is considered \textit{sufficient} if it is consistent with the provided claims and evidence while avoiding redundancy with other spans. 
This module provides span-level sufficiency scores that complement token-level confidence, enabling the system to revise both uncertain tokens and confidently produced but unsupported statements. 
We explore two complementary approaches.

\textbf{Chain-of-Thought Prompting.}  
We construct few-shot prompts for large language models to explicitly reason about sufficiency. 
Given a candidate span $s$ and its supporting arguments $\{(c_i, E_i)\}$, the model is prompted to determine whether $s$ is fully supported, insufficient, or redundant. 
This prompting strategy provides interpretable rationales and flexible adaptation across domains, though it may introduce variability depending on model scale and prompting quality.

\textbf{Supervised Sufficiency Classifier.}  
In parallel, we develop a RoBERTa-based classifier to provide stable sufficiency signals. 
For each span $s$, claim $c_i$, and evidence $E_i$, the classifier predicts a binary sufficiency label:
\begin{equation}
\mathcal{C}_\theta(s, c_i, E_i) = \sigma\left( f_\theta \left( [\![s; c_i; E_i]\!] \right) \right),
\end{equation}
where $f_\theta$ is the encoder and $[\![\cdot]\!]$ denotes concatenation with segment indicators. 
Training data are constructed via controlled perturbations (hallucinated, contradictory, or unsupported variants) to ensure coverage of common insufficiency types. 
During inference, the classifier provides span-level sufficiency scores $\text{Suff}(s_i)$ that are integrated into the corruption distribution, complementing low-confidence token masking.

\subsection{Diffusion Generation Framework}
\label{diffusion}

Inspired by recent advances in masked diffusion language models \citep{DBLP:journals/corr/abs-2502-09992}, we adopt a non-autoregressive, iterative generation paradigm for argument summarization. Unlike autoregressive decoders, which generate outputs token-by-token in a fixed order and struggle with revising earlier errors, our diffusion-based approach refines summaries through repeated masked denoising steps. This iterative design allows for flexible editing, span-level abstraction, and improved controllability—characteristics well-aligned with the requirements of factual and concise argument summarization.

Given an input set $\mathcal{X} = \{(c_i, E_i)\}_{i=1}^N$ consisting of claims and their supporting evidence, the goal is to generate a coherent and faithful summary $\hat{S}$. To achieve this, we initialize with a fully masked summary $\tilde{S}_T$ and iteratively denoise it using a learned reverse process. At each timestep $t$, we sample a cleaner intermediate summary $\tilde{S}_{t-1}$ as follows:

\begin{equation}
    \tilde{S}_{t-1} \sim q_{\theta}(\tilde{S}_{t-1} \mid \tilde{S}_t, \mathcal{X}), \quad t = T, ..., 1
\end{equation}

This process continues until convergence, yielding the final summary $\hat{S} = \tilde{S}_0$.

\paragraph{Training Objective.}
Following \citet{DBLP:journals/corr/abs-2502-09992}, the model is trained via a denoising score matching objective. Specifically, given a reference summary $\hat{S}$, we randomly corrupt a subset of tokens $\mathcal{M} \subset \hat{S}$ and optimize the model to reconstruct the original tokens conditioned on the unmasked context and input arguments:

\begin{equation}
    \mathcal{L}_{\text{diff}} = - \mathbb{E}_{\mathcal{M}} \sum_{t \in \mathcal{M}} \log p_{\theta}(\hat{S}_t \mid \hat{S}_{\backslash \mathcal{M}}, \mathcal{X})
\end{equation}

This objective encourages the model to generate contextually appropriate and semantically grounded content. Importantly, masking during training remains random to ensure broad denoising capability, while targeted refinement is deferred to inference.

\subsection{Inference-Time Sufficiency Guided Masking}
\label{masking}

In line with recent advances in diffusion-based text generation, Arg-LLaDA adopts an inference-time selective masking strategy to refine summaries across iterative denoising steps. 
Instead of relying solely on random corruption, our framework integrates two complementary signals—token-level confidence and span-level sufficiency. 
This combined design allows the model to address both locally uncertain expressions and confidently generated but unsupported statements, thereby achieving more faithful and concise summaries.

\textbf{Token-level low-confidence remasking.} 
At each inference step, the model produces probability distributions over candidate tokens. 
We compute token confidence as the maximum probability assigned to any candidate token. 
Tokens with confidence below a threshold $\tau$ are deemed unreliable, and the lowest-$r$ proportion of such tokens are selected for resampling. 
This mechanism ensures that local corrections are directed toward genuinely uncertain regions while avoiding unnecessary revisions of fluent and stable tokens. 
Following prior work on diffusion-based generation \citep{DBLP:journals/corr/abs-2502-09992}, we apply confidence-based remasking only at inference, not training. 
This separation ensures that the diffusion backbone learns a broad, task-agnostic denoising capability, while inference-time confidence signals specialize revision to context-specific uncertainties.

\textbf{Sufficiency-guided span masking.}  
Low-confidence detection alone cannot address confidently incorrect content or omissions. 
To mitigate this, we integrate an external sufficiency module $\mathcal{C}_\phi$, which evaluates whether candidate summary spans are adequately supported by the input claims and evidence. 
Spans are segmented primarily at the sentence level, aligned with punctuation and discourse boundaries. 
For compound sentences, we further split into clause-level units to obtain semantically coherent argumentative statements. 
Each span is assigned a sufficiency score or label, and tokens within spans flagged as insufficient are projected into the corruption distribution.

\textbf{Combined corruption distribution.}  
Both signals are unified into a single masking distribution:
\begin{equation}
p_{\text{mask}}(i) \propto (1 - \text{Conf}(s_i)) + \gamma \cdot (1 - \text{Suff}(s_i)),
\end{equation}
where $\text{Conf}(s_i)$ is the token-level confidence, $\text{Suff}(s_i)$ the span-level sufficiency assigned to the token’s enclosing span, and $\gamma$ a balancing coefficient. 
At each denoising step, we sample tokens for resampling according to this distribution. 

\subsection{Training and Implementation Details}
\label{training}

We adopt a two-stage training strategy for \textbf{Arg-LLaDA}, comprising diffusion-based denoising and sufficiency classification. 
Importantly, sufficiency guidance is introduced only at inference time: training uses diverse random masking to teach general denoising skills, while inference selectively leverages sufficiency signals for targeted refinement. 
This separation avoids overfitting to noisy sufficiency pseudo-labels and maintains generalization.

\paragraph{Diffusion Backbone Fine-tuning.}
We initialize the diffusion model with the LLaDA checkpoint \citep{DBLP:journals/corr/abs-2502-09992}, pretrained on large-scale web corpora with masked diffusion objectives. 
Its architecture and noise scheduling are retained. 
For fine-tuning, we supervise the model on domain-specific argument summarization data. 
Given input tuples of claims and evidence with reference summaries, we apply stochastic masking (30\% ratio). 
The model is then optimized to denoise the masked summary using the diffusion objective (Section~\ref{diffusion}).

\paragraph{Sufficiency Classifier.}
We train a RoBERTa-base encoder \citep{DBLP:journals/corr/abs-1907-11692} to predict span-level sufficiency labels. 
Each training instance consists of a sentence-level span \(s\), its associated claim \(c_i\), and supporting evidence \(E_i\). 
Labels are automatically generated through controlled perturbations, including hallucinations, contradictions, and omissions. 
The classifier is trained with binary cross-entropy loss and applied only during inference to guide selective span masking.

\paragraph{Optimization and Setup.}
We use the AdamW optimizer with linear learning rate decay. 
Batch size is set to 32, with early stopping based on validation BLEURT score. 
All experiments are conducted on two NVIDIA A100 80GB GPUs with FP16 precision.
\section{Experimental Setup}

Broadly speaking, we aim to investigate the effectiveness of our proposed \textbf{Arg‑LLaDA} framework in addressing the limitations of current argument summarization methods. Specifically, we pose the following research questions:
\emph{(i)} Does \textbf{Arg‑LLaDA} improve the overall quality of argument summaries compared to standard one-pass or prompt-based methods?
\emph{(ii)} To what extent does the sufficiency-based span diagnosis contribute to reducing redundancy and enhancing faithfulness?
\emph{(iii)} What is the impact of different sufficiency checking modules?

\paragraph{Baselines.} We compare \textbf{Arg‑LLaDA} against ten state-of-the-art baselines, categorized into three groups:
\emph{(i)} Zero-shot proprietary LLMs, including GPT-4 \cite{DBLP:journals/corr/abs-2303-08774}, GPT-o3 \footnote{https://openai.com/index/introducing-o3-and-o4-mini/}, Gemini 2.5 \citep{DBLP:journals/corr/abs-2312-11805}, Claude 3.7 \footnote{https://www.anthropic.com/news/claude-3-7-sonnet}, and DeepSeek-R1 \citep{DBLP:journals/corr/abs-2501-12948}, which are closed-source commercial models accessed via API without any task-specific adaptation;
\emph{(ii)} Zero-shot open-source LLMs, such as LLaMA 3 \citep{DBLP:journals/corr/abs-2407-21783}, LLaDA \citep{DBLP:journals/corr/abs-2502-09992}, Mistral \citep{DBLP:journals/corr/abs-2310-06825}, Gemma \citep{DBLP:journals/corr/abs-2408-00118} and FlanT5 \citep{DBLP:journals/jmlr/ChungHLZTFL00BW24}, which are publicly released and evaluated in their instruction-tuned form without additional training;
\emph{(iii)} Full-shot open-source LLMs, which are further fine-tuned on benchmark datasets to adapt them explicitly to the argument summarization task. 

\paragraph{Dataset.} The ArgKP dataset consists of approximately 27k labeled argument–key point pairs covering 31 diverse debate topics, with each pair annotated for stance polarity \citep{DBLP:conf/acl/Bar-HaimEFKLS20,DBLP:conf/emnlp/Bar-HaimKEFLS20}. Following prior work, we partition the dataset into training and test sets in a 28:3 ratio while preserving topic distribution. In addition, we employ and extand the ASE dataset, which provides a comprehensive end-to-end benchmark for argument \citep{DBLP:conf/acl/0074WSBMZZWHLN24} summarization. It spans the full pipeline—from claim and evidence detection to the generation and evaluation of debate-ready summaries—enabling holistic assessment of argument generation systems. 

\begin{table*}[t]
    \centering    
    \begin{adjustbox}{width=\textwidth}
    \begin{tabular}{clcccccccccc}
        \toprule
        & \multirow{2}{*}{\textbf{Methods}} & \multicolumn{5}{c}{\textit{ASE}} & \multicolumn{5}{c}{\textit{ArgKP}} \\
        \cmidrule(lr){3-7}  \cmidrule(lr){8-12}
        & & R-1 & R-2 & R-L & BLE & BER & R-1 & R-2 & R-L & BLE & BER \\ 
        \midrule
        \multirow{10}{*}{\rotatebox[origin=c]{90}{\textbf{Zero-shot}}} 
        & \textit{GPT-4} & $0.314$ & $0.091$ & $0.302$ & $0.512$ & $0.703$ & $0.402$ & $0.192$ & $0.374$ & \underline{$0.656$} & $0.809$ \\
        & \textit{GPT-o3} & $0.333$ & $0.105$ & $0.311$ & $0.566$ & $0.722$ & $0.453$ & $0.233$ & $0.419$ & $0.619$ & $0.811$ \\
        & \textit{Gemini 2.5} & $0.341$ & $0.128$ & $0.337$ & $0.577$ & \underline{$0.744$} & $0.468$ & $0.245$ & $0.436$ & $0.637$ & $0.823$ \\
        & \textit{Claude} & $0.334$ & $0.119$ & $0.326$ & $0.568$ & $0.740$ & $0.454$ & $0.229$ & $0.418$ & $0.626$ & $0.820$ \\
        & \textit{DeepSeek-R1} & $0.361$ & $0.137$ & $0.342$ & $\textbf{0.587}$ & $\textbf{0.746}$  & $0.475$ & $0.257$ & $0.441$ & $0.601$ & $0.769$ \\
        & \textit{LLaMA 3.1} & $0.328$ & $0.097$ & $0.304$ & $0.524$ & $0.721$ & $0.443$ & $0.215$ & $0.406$ & $0.614$ & $0.818$\\
        & \textit{LLaDA} & $0.332$ & $0.105$ & $0.312$ & $0.545$ & $0.721$ & $0.455$ & $0.231$ & $0.410$ & $0.618$ & $0.921$ \\
        & \textit{Mistral} & $0.291$ & $0.083$ & $0.277$ & $0.503$ & $0.683$ & $0.344$ & $0.228$ & $0.305$ & $0.512$ & $0.781$ \\
        & \textit{Gemma} & $0.282$ & $0.081$ & $0.276$ & $0.544$ & $0.714$ & $0.457$ & $0.233$ & $0.422$ & $0.629$ & $0.822$ \\
        & \textit{FlanT5} & $0.266$ & $0.076$ & $0.235$ & $0.492$ & $0.667$ & $0.342$ & $0.197$ & $0.313$ & $0.402$ & $0.754$  \\
        \midrule
        \multirow{5}{*}{\rotatebox[origin=c]{90}{\textbf{Fine-tuning}}} 
        & \textit{LLaMA 3.1} & $0.353$ & $0.122$ & $0.338$ & $0.571$ & $0.728$ & $0.493$ & $0.266$ & $0.421$ & $0.635$ & $0.822$\\
        & \textit{LLaDA} & $0.355$ & $0.131$ & $0.347$ & $0.575$ & $0.738$ & $0.501$ & $0.267$ & $0.428$ & $0.629$ & $0.813$ \\
        & \textit{Mistral} & $0.325$ & $0.099$ & $0.307$ & $0.533$ & $0.726$ & $0.382$ & $0.183$ & $0.347$ & $0.537$ & $0.784$ \\
        & \textit{Gemma} & $0.300$ & $0.099$ & $0.291$ & $0.532$ & $0.710$ &  \underline{$0.511$} &  \underline{$0.269$} & $\textbf{0.455}$ & $0.643$ & \underline{$0.830$} \\
        & \textit{FlanT5} & $0.290$ & $0.085$ & $0.281$ & $0.503$ & $0.684$ & $0.404$ & $0.203$ & $0.382$ & $0.551$ & $0.797$ \\
        \midrule
        & \textbf{Arg-LLaDA} & $\textbf{0.363}$ & $\textbf{0.138}$ & $\textbf{0.349}$ &  \underline{$0.580$} & $0.741$ & $\textbf{0.525}$ & $\textbf{0.271}$ & \underline{$0.453$} & $\textbf{0.658}$ & $\textbf{0.834}$ \\
        \bottomrule
    \end{tabular}
    \end{adjustbox}
    \caption{Comparison of \textbf{Arg‑LLaDA} against zero-shot and fine-tuned LLMs on two argument summarization datasets. We report ROUGE (R-1, R-2, R-L), BLEURT, and BERTScore. The best and second-best results are respectively \textbf{bolded} and \underline{underlined}.}
    \label{tab:arg_summary_results}
\end{table*}

\paragraph{Metrics.} We report \textbf{ROUGE} \citep{lin2004rouge}, which common used in text summarization. While ROUGE captures surface-level similarity, it may overlook semantic equivalence in cases of lexical variation. To do so, we consider \textbf{BERTScore} \citep{DBLP:conf/iclr/ZhangKWWA20}  and \textbf{BLEURT} \citep{DBLP:conf/acl/SellamDP20} evaluates summary quality based on contextualized token embeddings obtained from a pre-trained BERT model. They better account for semantic similarity than ROUGE and are less sensitive to exact word matching.

\paragraph{Human Evaluation.} While automatic metrics provide efficient ways to evaluate summarization quality, they are known to exhibit several limitations—particularly in the context of argument summarization. These metrics primarily rely on surface-level lexical or semantic overlap with reference summaries, and thus often fail to capture deeper aspects such as logical structure, factual grounding, and the nuanced coverage of multiple argumentative facets. Moreover, they may assign high scores to outputs that are fluent but contain hallucinations or overlook important viewpoints.

To address these shortcomings, we complement our automatic evaluation with a human assessment focused on three essential dimensions: \textbf{Coverage}, \textbf{Faithfulness}, and \textbf{Conciseness} \citep{DBLP:journals/corr/abs-2503-00847}. These criteria reflect core desiderata of argumentative summaries—namely, the ability to accurately capture all salient perspectives, avoid unsupported or fabricated content, and maintain clarity without redundancy.

Each summary is evaluated by three trained annotators with backgrounds in argument mining or discourse analysis. Given the input claims and evidence, annotators rate the summary using a 5-point Likert scale (1 = very poor, 5 = excellent) according to:
\begin{itemize}
    \item \textbf{Coverage}: Does the summary reflect all relevant and diverse viewpoints from the input?
    \item \textbf{Faithfulness}: Are the summarized statements well-grounded in the original claims and evidence, without introducing hallucinated or contradictory content?
    \item \textbf{Conciseness}: Is the summary free from redundancy and excessive verbosity, while still conveying the complete argumentative structure?
\end{itemize}

All outputs are anonymized and randomly ordered to mitigate evaluator bias. We report the average scores across all annotators, and assess inter-annotator agreement via Krippendorff’s alpha. This evaluation offers a more comprehensive view of summary quality, complementing automatic scores with expert human judgment. Detail could be find at Appendix \ref{appendix:human}.
\section{Results and Analysis}
\label{sec:exp-results}

To evaluate the effectiveness of our proposed \textbf{Arg‑LLaDA} framework, we revisit the three core research questions posed earlier. 

\paragraph{Arg‑LLaDA significantly outperforms both zero-shot and fine-tuned large language models (LLMs) in generating high-quality argument summaries.}

As shown in Table~\ref{tab:arg_summary_results}), Arg‑LLaDA achieves the highest ROUGE-1 ($0.525$), ROUGE-2 ($0.271$), and ROUGE-L ($0.453$) on the \textit{ArgKP} dataset, outperforming both fine-tuned (\textit{Gemma}, \textit{LLaDA}) and zero-shot (\textit{Gemini 2.5}, \textit{DeepSeek-R1}) baselines. These results are consistent across the \textit{ASE} dataset, where Arg‑LLaDA also leads in ROUGE-L ($0.349$), BLEURT ($0.580$), and BERTScore ($0.741$), indicating strong lexical and semantic overlap with human-written references. Compared to its closest competitors—\textit{Gemini 2.5} and \textit{DeepSeek-R1}—Arg‑LLaDA shows modest but consistent improvements across all dimensions, particularly in BLEURT and BERTScore, which are more sensitive to semantic adequacy and contextual fidelity.

\begin{figure}
    \centering
    \includegraphics[scale=0.3]{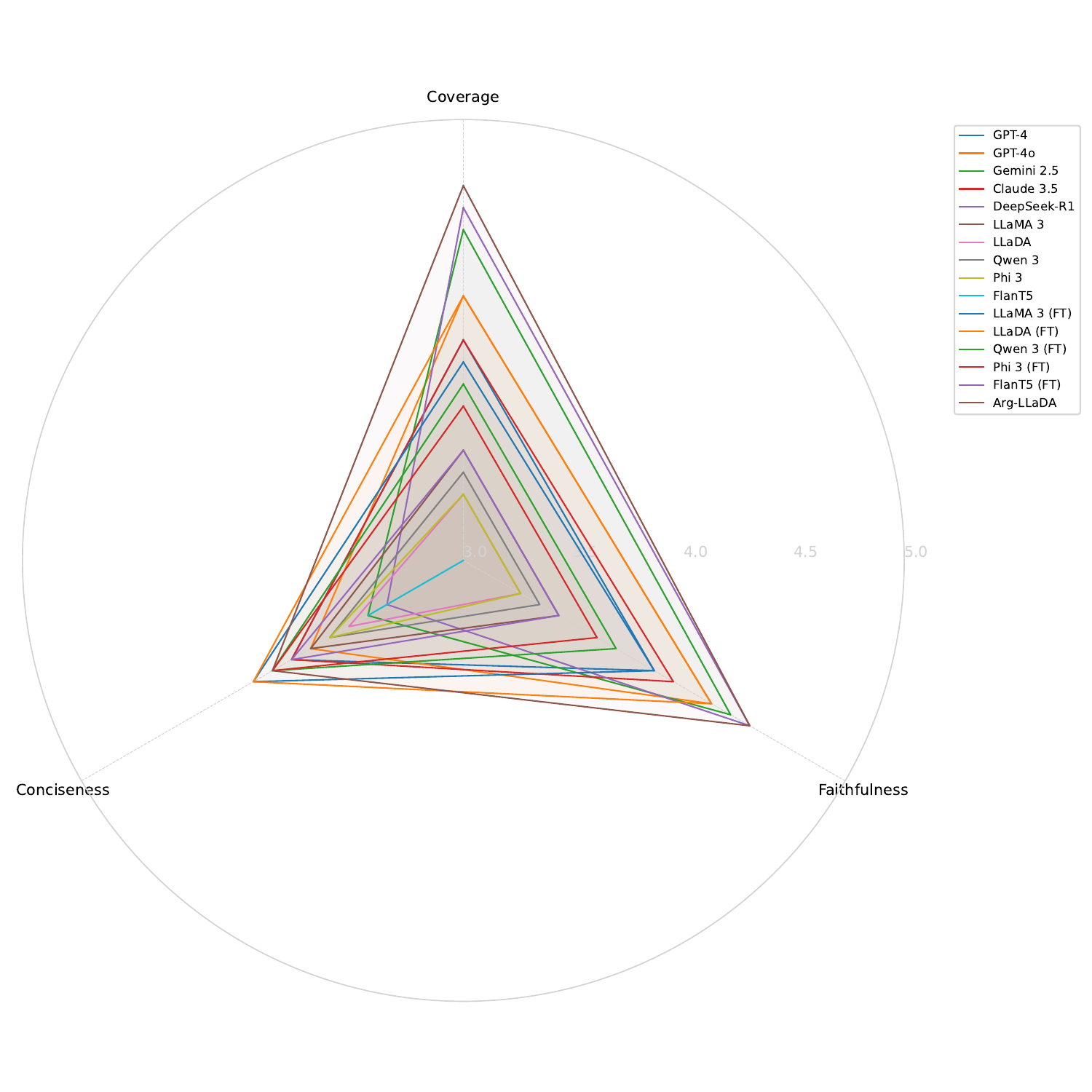}
    \caption{Human evaluation results across three dimensions, using a 5-point Likert scale. }
    \label{fig:human_eval}
\end{figure}

Human evaluation (Figure~\ref{fig:human_eval}) further corroborates these findings. Arg‑LLaDA receives the highest average scores across all three dimensions: Coverage ($4.7$), Faithfulness ($4.6$), and Conciseness ($4.4$). These results suggest that Arg‑LLaDA not only captures a broader spectrum of argumentative content but also avoids common issues such as hallucination or redundancy. In contrast, while models like \textit{Gemini 2.5} ($4.5/4.4/4.2$) and \textit{DeepSeek-R1} ($4.4/4.3/4.1$) exhibit strong performance, they tend to generate longer outputs that introduce occasional repetition or irrelevant detail, resulting in slightly lower conciseness scores.

\paragraph{The integration of sufficiency-based span diagnosis plays a pivotal role in improving both the conciseness and faithfulness of generated summaries.}

As shown in Figure~\ref{fig:human_eval}, Arg‑LLaDA outperforms all baselines in the Faithfulness ($4.6$) and Conciseness ($4.4$) dimensions. These improvements can be attributed to the model’s explicit mechanism for identifying and filtering spans that lack sufficient argumentative support before generation. Compared to its fine-tuned counterparts (e.g., \textit{Gemma}, $4.3/4.2$; \textit{LLaDA}, $4.2/4.1$), Arg‑LLaDA demonstrates superior control over hallucinated or redundant content. This is particularly evident when contrasting with zero-shot models like \textit{Claude} or \textit{Qwen 3}, whose summaries frequently exhibit ungrounded generalizations or verbose phrasing, as reflected in lower average ratings for faithfulness and conciseness (both $<4.0$). Importantly, our human evaluations demonstrate not only absolute performance gains but also high reliability and alignment with automatic metrics. Across five annotators, Krippendorff’s~$\alpha$ \citep{krippendorff2011computing} reaches substantial agreement for all three dimensions (Coverage~$0.69$, Faithfulness~$0.74$, Conciseness~$0.66$). Furthermore, Soft scores exhibit strong monotonic trends with human judgments, confirming that our automatic metrics meaningfully reflect annotator preferences.

\begin{table}[htbp]
    \centering
    \resizebox{0.48\textwidth}{!}{
        \begin{tabular}{lccc}
            \toprule
            \textbf{Iterations} & \textbf{R-L} & \textbf{BLEURT} & \textbf{BERTScore} \\
            \midrule
            0 & $0.429$ & $0.629$ & $0.814$ \\
            1 & $0.448$ & $0.650$ & $0.829$ \\
            2 & $0.451$ & $0.657$ & $0.831$ \\
            3 & $0.453$ & $0.658$ & $0.834$ \\
            \bottomrule
        \end{tabular}
    }
    \caption{Ablation study: varying the number of iterative refinement steps.}
    \label{tab:ablation_iterations}
\end{table}

\paragraph{The design and integration of the sufficiency checking module critically affect final summarization performance, both in lexical overlap and semantic adequacy.}

As shown in Table~\ref{tab:ablation_sufficiency}, removing the sufficiency diagnosis module (“No Diagnosis”) results in a significant drop across all evaluation metrics—ROUGE-L drops from $0.453$ to $0.428$, BLEURT from $0.658$ to $0.629$, and BERTScore from $0.834$ to $0.813$. This indicates that without explicit sufficiency filtering, the generation becomes more verbose and includes unsupported content, thereby reducing both faithfulness and conciseness.

\begin{table}[htbp]
    \centering
    \resizebox{0.48\textwidth}{!}{
        \begin{tabular}{lccc}
            \toprule
            \textbf{Sufficiency Module} & \textbf{R-L} & \textbf{BLEURT} & \textbf{BERTScore} \\
            \midrule
            No Diagnosis & $0.428$ & $0.629$ & $0.813$ \\
            CoT Prompting Only & $0.451$ & $0.649$ & $0.827$ \\
            Classifier Only & $0.447$ & $0.645$ & $0.823$ \\
            Combine & $0.453$ & $0.658$ & $0.834$ \\
            \bottomrule
        \end{tabular}
    }
    \caption{Ablation study: comparing sufficiency diagnosis strategies.}
    \label{tab:ablation_sufficiency}
\end{table}

When comparing alternative sufficiency strategies, we observe that simple classifier-based diagnosis yields moderate improvements over the no-diagnosis baseline ($+0.019$ in ROUGE-L), while CoT-style prompting achieves slightly better results. However, our proposed combined approach, which integrates classifier predictions with chain-of-thought (CoT) rationale prompting, consistently achieves the best results across all metrics—demonstrating a synergistic effect when both implicit (reasoning-driven) and explicit (discriminative) sufficiency signals are fused.

\section{Conclusion}

In this work, we introduced \textbf{Arg‑LLaDA}, a novel large language diffusion framework for argument summarization that integrates sufficiency-guided remasking and iterative regeneration. Our method addresses key limitations of existing approaches by enabling controllable refinement of summaries through a flexible masking controller and sufficiency diagnosis module. Empirical results across two benchmark datasets demonstrate that Arg‑LLaDA consistently outperforms state-of-the-art baselines. These findings highlight the potential of diffusion-based self-refinement in enhancing the factuality and structure of argumentative summaries.

\section*{Limitations}

Despite the promising results of \textbf{Arg‑LLaDA}, several limitations warrant discussion. First, although we trained annotators with backgrounds in argument mining and provided clear evaluation guidelines, human judgments remain inherently subjective. Variations in interpretation or attention to detail may affect the consistency of annotation, especially for nuanced aspects like sufficiency and redundancy. Second, while our datasets cover a wide range of debate topics, some of these topics may contain controversial or sensitive content. But such content is not the focus of this work.

\bibliography{custom}

\appendix


\section{Instruction for Zero-shot Inference.}

We provide the instruction we used to prompt baseline model to complete the summary task on both dataset:


\textit{You are a debate writing assistant. Based on the structured information provided below, generate a \textbf{concise and persuasive debate speech} in support of the motion: \textbf{``[Insert Debate Topic]''}.}

\textit{The speech should follow this structure:}

\begin{itemize}
  \item \textit{\textbf{Introduction}: Clearly state the affirmative stance and why the topic matters.}
  \item \textit{\textbf{Body}: Present each supporting claim, explain its reasoning, and integrate the evidence in a natural, narrative style. You do not need to use all the evidence—just enough to make each point persuasive.}
  \item \textit{\textbf{Conclusion}: Summarize the main points and restate why the position is valid and important.}
\end{itemize}

\textit{Use the following input fields to guide the speech:}


\textit{Guidelines:}
\begin{itemize}
    \item \textit{Select evidence selectively—strong, relevant examples are enough.}
    \item \textit{Write in fluent, professional, and persuasive English.}
    \item \textit{Output should be in paragraph format, suitable for public speaking.}
    \item \textit{Avoid exaggeration; keep the reasoning grounded in the given claims and evidence.}
\end{itemize}

\section{Human Evaluation Guidelines}
\label{appendix:human}

To complement automatic metrics with more semantically grounded assessments, we employ a controlled human evaluation procedure following established practices in argument summarization. This appendix provides the full annotation guidelines, evaluator instructions, and quality-control measures used in our study.

\subsection{Annotator Background and Training}

All summaries were evaluated by \textbf{three trained annotators} with strong English proficiency, i.e. \textbf{native English speakers} or hold a \textbf{Master’s degree from an English-speaking country}.

Annotators first received a \textbf{guideline document} detailing the evaluation criteria and were asked to review example summaries with explanations. We then conducted a \textbf{calibration round} of 15 practice examples to harmonize interpretation of the rubric; disagreements were discussed and resolved before the official annotation began.

\subsection{Evaluation Setup}

For each instance, annotators were presented with:
\begin{itemize}
    \item the \textbf{input claim},
    \item the set of \textbf{retrieved evidence spans}, and
    \item a \textbf{randomly ordered, anonymized list of model-generated summaries}
\end{itemize}

to prevent ordering bias and model-identification effects.

Annotators rated each summary independently using a \textbf{5-point Likert scale} (1 = very poor, 5 = excellent) along three dimensions defined below.

\subsection{Rating Rubric}
\paragraph{Coverage (1–5 Scale)}
\begin{itemize}
    \item \textbf{5 – Excellent}: Captures all major arguments and viewpoints with no omissions.
    \item \textbf{4 – Good}: Covers most key points, with only minor or nuanced elements missing.
    \item \textbf{3 – Fair}: Conveys the main idea but omits several relevant viewpoints or supporting arguments.
    \item \textbf{2 – Poor}: Misses many important aspects; only captures a small portion of the argument space.
    \item \textbf{1 – Very Poor}: Fails to reflect the input’s arguments; largely incomplete or irrelevant.
\end{itemize}

\paragraph{Faithfulness (1–5 Scale)}
\begin{itemize}
    \item \textbf{5 – Excellent}: All statements are fully grounded in the input; no hallucinations or distortions.
    \item \textbf{4 – Good}: Mostly faithful with only minor unsupported phrasing or slight overgeneralization.
    \item \textbf{3 – Fair}: Generally accurate but contains noticeable paraphrasing errors or weakly supported claims.
    \item \textbf{2 – Poor}: Includes several unsupported, incorrect, or contradictory statements.
    \item \textbf{1 – Very Poor}: Largely unfaithful; contains substantial hallucinations or fabrications.
\end{itemize}

\paragraph{Conciseness (1–5 Scale)}
\begin{itemize}
    \item \textbf{5 – Excellent}: Succinct and clear; no redundancy; conveys the full argument with minimal wording.
    \item \textbf{4 – Good}: Mostly concise; contains only minor repetition of claim or slightly verbose phrasing.
    \item \textbf{3 – Fair}: Understandable but includes some redundancy or unnecessary detail.
    \item \textbf{2 – Poor}: Noticeably verbose, repetitive, or cluttered; key content is diluted.
    \item \textbf{1 – Very Poor}: Highly redundant or rambling; lacks clarity and wastes space.
\end{itemize}

\section{Detailed Experiment Result}

\begin{table*}[t]
\centering
\begin{adjustbox}{width=\textwidth}
\small
\begin{tabular}{lcccccccc}
\toprule
 & GPT-4 & GPT-o3 & Gemini 2.5 & Claude & DeepSeek-R1 & LLaMA 3.1 & LLaDA & Mistral \\
\midrule
$sP_{\text{BLEURT}}$  & 0.665 & 0.638 & 0.650 & 0.645 & 0.588 & 0.620 & 0.633 & 0.525 \\
$sR_{\text{BLEURT}}$  & 0.647 & 0.602 & 0.626 & 0.607 & 0.615 & 0.608 & 0.604 & 0.500 \\
$sF1_{\text{BLEURT}}$ & 0.656 & 0.619 & 0.637 & 0.626 & 0.601 & 0.614 & 0.618 & 0.512 \\
\midrule
\addlinespace[2pt]
$sP_{\text{BERT}}$  & 0.823 & 0.825 & 0.842 & 0.838 & 0.748 & 0.819 & 0.932 & 0.767 \\
$sR_{\text{BERT}}$  & 0.795 & 0.799 & 0.804 & 0.803 & 0.791 & 0.815 & 0.910 & 0.797 \\
$sF1_{\text{BERT}}$ & 0.809 & 0.811 & 0.823 & 0.820 & 0.769 & 0.818 & 0.921 & 0.781 \\
\midrule
 & Gemma & FlanT5 & LLaMA 3.1 (FT) & LLaDA (FT) & Mistral (FT) & Gemma (FT) & FlanT5 (FT) & Arg-LLaDA \\
\midrule
$sP_{\text{BLEURT}}$  & 0.618 & 0.415 & 0.640 & 0.643 & 0.545 & 0.652 & 0.545 & 0.671 \\
$sR_{\text{BLEURT}}$  & 0.642 & 0.390 & 0.631 & 0.616 & 0.530 & 0.637 & 0.559 & 0.645 \\
$sF1_{\text{BLEURT}}$ & 0.629 & 0.402 & 0.635 & 0.629 & 0.537 & 0.643 & 0.551 & 0.658 \\
\midrule
\addlinespace[2pt]
$sP_{\text{BERT}}$  & 0.823 & 0.760 & 0.835 & 0.834 & 0.810 & 0.841 & 0.805 & 0.854 \\
$sR_{\text{BERT}}$  & 0.817 & 0.748 & 0.808 & 0.794 & 0.764 & 0.820 & 0.790 & 0.815 \\
$sF1_{\text{BERT}}$ & 0.822 & 0.754 & 0.822 & 0.813 & 0.784 & 0.830 & 0.797 & 0.834 \\
\bottomrule
\end{tabular}
\end{adjustbox}

\caption{
\textbf{Detailed Soft-Score Evaluation on ArgKP.}  
We report Soft-Precision, Soft-Recall, and Soft-F1 using two semantic similarity functions (BLEURT and BERTScore).  
}
\label{tab:argkp_softscores_full}
\end{table*}

\end{document}